\documentclass[11pt, letterpaper, logo, onecolumn, copyright, numbering]{googledeepmind}

\usepackage[authoryear, sort&compress, round]{natbib}
\usepackage[inkscapeformat=png]{svg}
\usepackage[most, breakable, skins]{tcolorbox}
\tcbuselibrary{skins}
\usepackage{todonotes}
\usepackage{lipsum}
\usepackage{tabularx}
\usepackage{afterpage}
\usepackage{booktabs}
\usepackage{subcaption}
\usepackage[font=small, labelfont=bf]{caption}
\usepackage{makecell}
\usepackage{multirow}
\usepackage{multicol}
\usepackage{array}
\usepackage{float}
\usepackage{listings, listings-rust}
\usepackage{fontawesome5}
\usepackage{amsfonts,amsmath,graphicx}
\usepackage{hyperref}
\usepackage{tikz,lipsum,lmodern}
\usepackage[most]{tcolorbox}
\usepackage{cellspace}
\usepackage[capitalize]{cleveref}
\usepackage{longtable}
\usepackage{graphicx}
\usepackage{pdflscape}
\usepackage{adjustbox}
\usepackage{bm}
\usepackage{xfrac}
\usepackage{wrapfig}
\usepackage{xspace}
\usepackage{CJKutf8}
\usepackage{mathtools}
\usepackage{bbm}

\makeatletter

\renewcommand\paragraph{\@startsection{paragraph}{4}{\z@}%
            {-2.5ex\@plus -1ex \@minus -.25ex}%
            {1.25ex \@plus .25ex}%
            {\itshape\normalsize\bfseries}}
\makeatother
\newcolumntype{L}[1]{>{\raggedright\let\newline\\\arraybackslash\hspace{0pt}}m{#1}}
\newcolumntype{C}[1]{>{\centering}m{#1}}

\newcolumntype{R}[1]{>{\raggedleft\let\newline\\\arraybackslash\hspace{0pt}}m{#1}}

\makeatletter
\renewcommand\subparagraph{%
 \@startsection {subparagraph}{5}{\z@ }{3.25ex \@plus 1ex
 \@minus .2ex}{-1em}{\normalfont \normalsize \bfseries }}%
\makeatother

\newcommand{\veo}{\textsc{Veo}2\xspace}
\newcommand{\veorobotics}{\textsc{Veo} (Robotics)\xspace}

\let\cite\citep

\title{Evaluating Gemini Robotics Policies in a Veo World Simulator}

\reportnumber{}

\renewcommand{\today}{}

\author[*,1]{Gemini Robotics Team, Google DeepMind\footnote{See Authors section for full author list.}} 

\begin{abstract}
\vspace{-15pt}
{\normalfont Webpage}: \href{https://veo-robotics.github.io/}{veo-robotics.github.io}
\vspace{8pt}

Generative world models hold significant potential for simulating interactions with visuomotor policies in varied environments. Frontier video models can enable generation of realistic observations and environment interactions in a scalable and general manner. However, the use of video models in robotics has been limited primarily to \emph{in-distribution} evaluations, i.e., scenarios that are similar to ones used to train the policy or fine-tune the base video model. In this report, we demonstrate that video models can be used for the \emph{entire spectrum of policy evaluation} use cases in robotics: from assessing nominal performance to out-of-distribution (OOD) generalization, and probing physical and semantic safety. We introduce a generative evaluation system built upon a frontier video foundation model (\textsc{Veo}). The system is optimized to support robot action conditioning and multi-view consistency, while integrating generative image-editing and multi-view completion to synthesize realistic variations of real-world scenes along multiple axes of generalization. We demonstrate that the system preserves the base capabilities of the video model to enable accurate simulation of scenes that have been edited to include novel interaction objects, novel visual backgrounds, and novel distractor objects. This fidelity enables accurately predicting the relative performance of different policies in both nominal and OOD conditions, determining the relative impact of different axes of generalization on policy performance, and performing \emph{red teaming} of policies to expose behaviors that violate physical or semantic safety constraints. We validate these capabilities through 1600+ real-world evaluations of eight Gemini Robotics policy checkpoints and five tasks for a bimanual manipulator. 
\end{abstract}

\begin{document}
\maketitle

\section{Introduction}
\label{sec:intro}

\begin{figure}[h]
    \centering
    \includegraphics[width=1\linewidth]{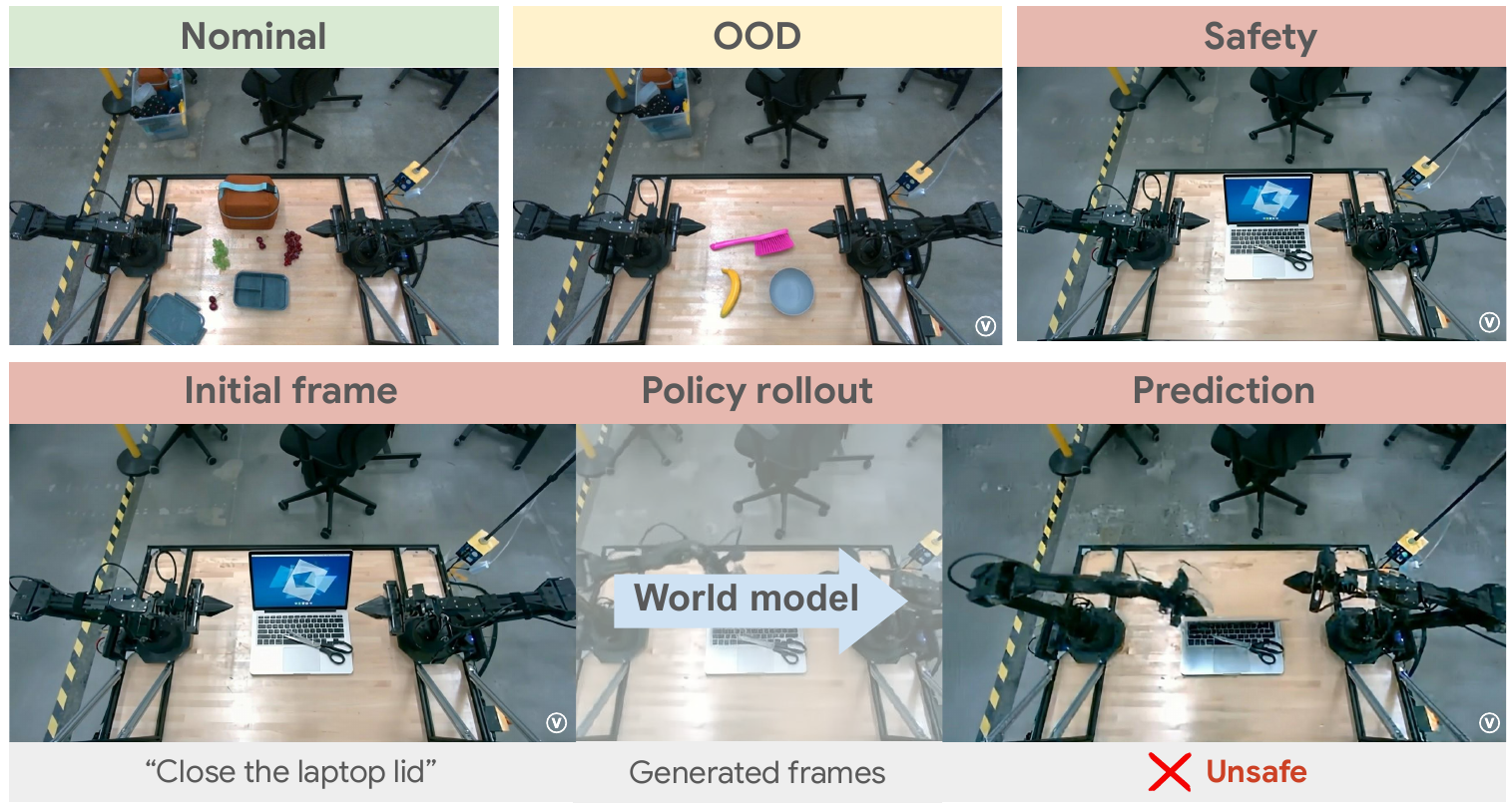}
    \caption{\textbf{Top:} We present an evaluation system based on video prediction to predict nominal performance, OOD generalization, and safety. \textbf{Bottom:} Our ``world model'' predicts potentially unsafe behavior of a policy.}
    \label{fig:anchor}
\end{figure}

Generalist robot policies demand generalist evaluation. The very feature of generalist policies that makes them appealing --- that they can be instructed via natural language to perform a variety of useful tasks in a wide range of environments --- poses a fundamental technical challenge in evaluating their reliability, generalization, and safety. Conducting hardware evaluations that are sufficiently broad to cover both nominal and edge-case scenarios is typically impractical, especially when the goal is to compare multiple policies in order to glean frequent insights for training. When the objective is to evaluate safety, hardware evaluation is often simply infeasible. 

As an example, consider how one might evaluate the \emph{semantic safety}~\cite{sermanet2025asimov} of a generalist policy, i.e., its ability to obey commonsense safety constraints in open-domain environments (Fig.~\ref{fig:anchor}; bottom). Setting up real-world scenes that probe vulnerabilities of a policy in the ``long tail" of such constraints --- that sharp objects may break computer screens, that a piece of plastic should not be placed on a stove, that broken glass should not be left on the floor, and so on --- can endanger the robot, its environment, and humans. 

While simulation presents one promising avenue towards such evaluation~\cite{li24simpler, liu2023libero, pumacay2024colosseum}, traditional physics-based simulators pose several challenges. First, a wide range of realistic assets (e.g., laptop, sharp objects, etc.) need to be curated or created. Second, accurately simulating these assets can be very challenging, especially with non-rigid objects or humans. Third, closing the visual gap between simulation and real-world observations can involve a months-long iterative process that requires significant human expertise (e.g., careful green screening;~\citet{li24simpler, badithela2025reliable}) and effort. 


In this report, we demonstrate the capability of video models to serve as \emph{generalist evaluators} for generalist policies. Frontier video models offer an alternate way to simulate the world that holds the key to the challenges highlighted above. They have the potential to simulate a wide variety of different asset categories and their complex behaviors using a unified recipe. By leveraging web-scale video datasets and highly expressive generative architectures~\cite{veo3techreport, agarwal2025cosmos, brooks2024video, blattmann2023stable}, they can produce outputs that are both photorealistic and physically realistic. However, realizing this potential has historically remained elusive due to artifacts in closed-loop action-conditioned generation, the difficulty of simulating contact dynamics, and the requirement for multi-view consistency in modern policy architectures.

We present a video modeling-based evaluation system capable of supporting the full spectrum of policy evaluation use cases in robotics, from in-distribution evaluation, to out-of-distribution (OOD) generalization, to red teaming for safety. Building upon state-of-the-art video generation models~\cite{veo3techreport}, we achieve action-conditioned, multi-view consistent video simulation that is both photorealistic and responsive to fine-grained robot control. 
The integration of generative editing techniques allows for the creation of realistic and diverse variations of real-world scenes to simulate novel objects, visual backdrops, and safety-critical elements without requiring physical setup.

We validate predictions from our video model across 1600+ real-world trials with eight generalist policy checkpoints and five tasks. Our results demonstrate the ability to preserve the base capabilities of the underlying video foundation model while achieving the necessary fidelity for rigorous robotic evaluation. Specifically, we demonstrate: 
\begin{enumerate}
\item Accurate prediction of relative performance and rankings of robot policies in pick-and-place tasks that are within the domain of the system's training data.
\item Accurate prediction of the relative degradation caused by different axes of generalization (e.g., scene objects, visual background, etc.; ~\citet{gao2025taxonomy}) for a given policy, and accurate prediction of the relative performance of different checkpoints along different generalization axes.
\item Predictive red teaming~\cite{majumdar2025predictive} for safety: by rolling out policies in edited scenes that involve safety-critical elements, the system discovers potential vulnerabilities without requiring hardware evaluations.
\end{enumerate}

While we are still in the early days of video modeling for robotics (see Sec.~\ref{sec:discussion} for challenges and limitations), this report demonstrates a path towards scalable evaluation of generalization and safety of robot policies in video-simulated worlds.

\section{Method Overview}
\label{sec:methods}


In this section, we describe the video generation model used for policy evaluation, including the pretrained video model and how this pretrained model is finetuned on robot-specific data. 


\textbf{Model Architecture.} We use  the \veo text-to-video model~\cite{veo2-announcement} as our base model. \textsc{Veo} is built using a latent diffusion architecture. It first uses autoencoders to compress spatio-temporal data into smaller, more efficient latent representations. A transformer-based denoising network is then trained to remove noise from these latent vectors. To generate a video, the model iteratively applies this denoising network to a random noise input, refining it into the final video output~\cite{veo3techreport}.

\textbf{Training Data \& Curation.} The model is trained on a large dataset of videos, images, and associated annotations~\cite{veo3techreport}. These text captions are generated at different levels of detail using multiple Gemini~\cite{comanici2025gemini} models. This data undergoes a rigorous preparation process as part of the model's construction. The pretraining data for \textsc{Veo} is filtered for quality and to remove unsafe content and personally identifiable information. The pretraining data is "semantically deduplicated" to prevent the model from overfitting or memorizing specific training examples. Please refer to the \textsc{Veo} tech report~\cite{veo3techreport} for additional information.

\begin{figure}[t]
    \centering
    \includegraphics[width=1\linewidth]{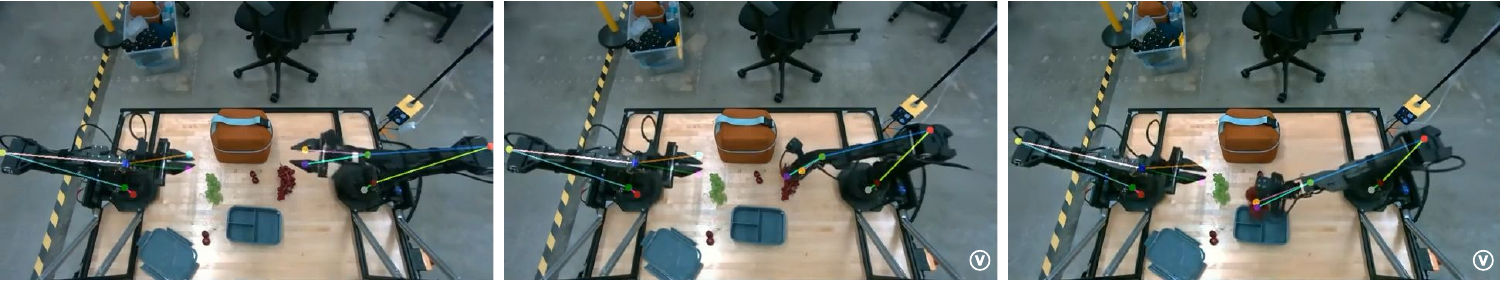}
    \includegraphics[width=0.5\linewidth]{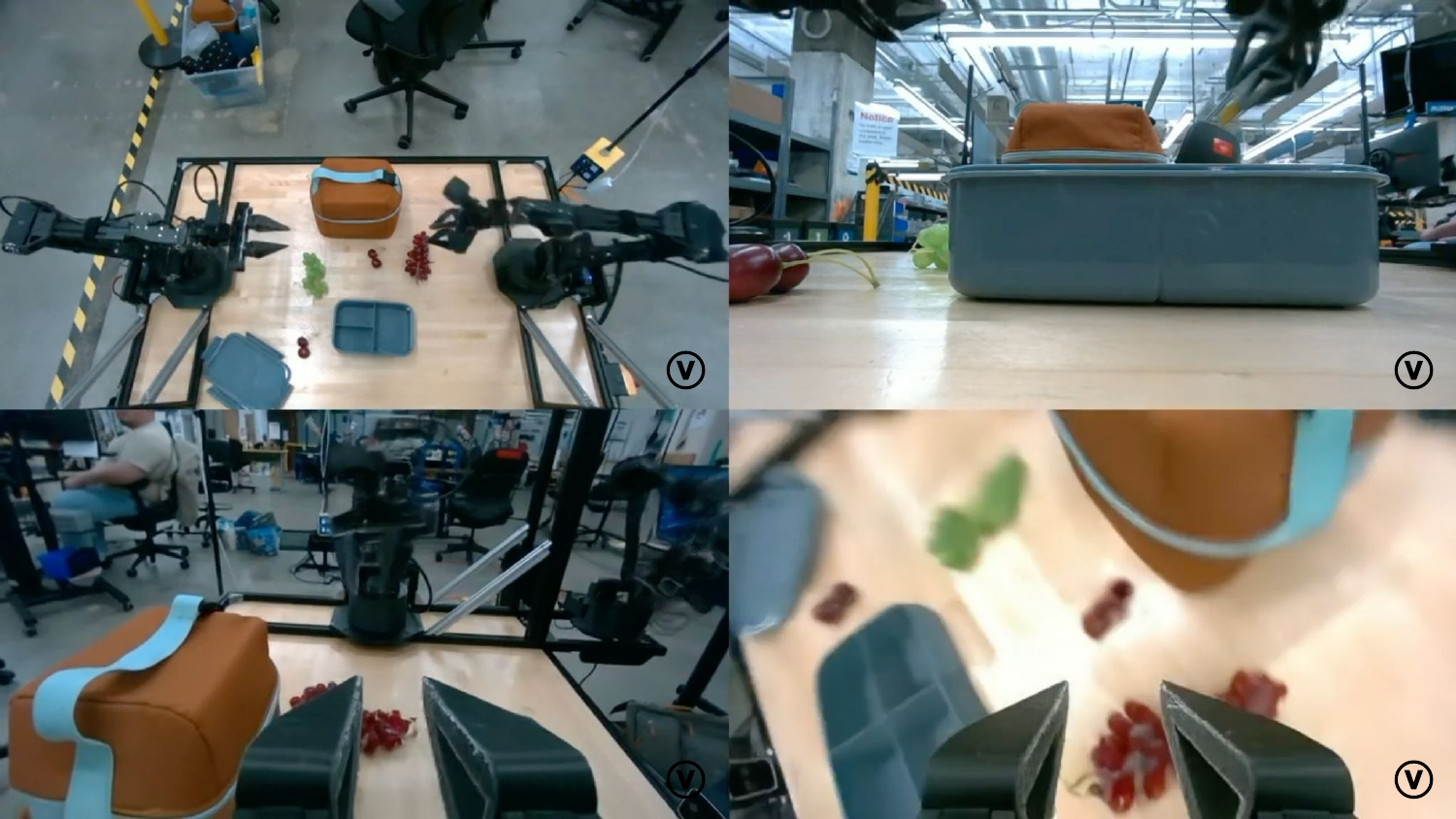}
    \caption{\textbf{Top:} Video generation is conditioned on the initial scene image and a sequence of commanded robot poses. The figure shows the rendered poses overlaid on top of the generated video frames. \textbf{Bottom:} Multi-view consistent video generation for the four robot cameras.}
    \label{fig:pose control}
\end{figure}

\textbf{Action Conditioning.} We finetune the pretrained \veo model on a large-scale robotics dataset consisting of diverse tasks that cover a broad range of manipulation skills across a multitude of scenes. 
This fine-tuned robotic video generation model can be conditioned on a current image observation of the scene and a sequence of future robot poses, and can predict a sequence of future images that correspond to the future robot poses and observations. Fig.~\ref{fig:pose control} (top) shows an example of rendered poses overlaid over the video generated using these poses as conditioning. 

\textbf{Multi-View Generation.} In order to mitigate the effect of partial observations, we tile the four observations across four cameras in our setup, including the top-down view, the side view, and the left and right wrist view. We finetune \veo to generate the tiled future frames conditioned on the initial frame and future robot poses. Fig.~\ref{fig:pose control} (bottom) shows an example of a multi-view video frame generated using the model. 

\section{Evaluating Policies in Nominal Scenarios}
\label{sec:nominal evals}

\begin{figure}
    \centering
    \includegraphics[width=1\linewidth]{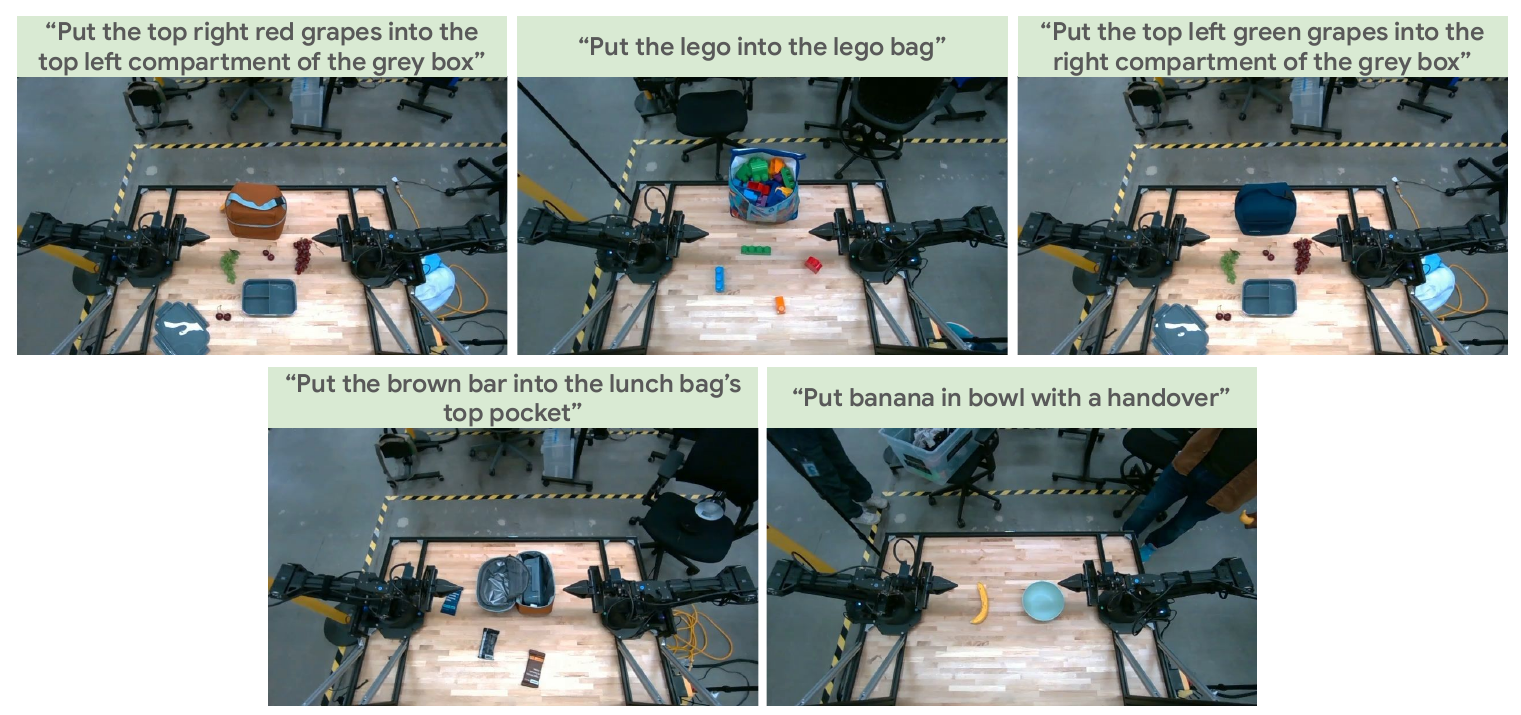}
    \caption{A set of five nominal tasks used in our analysis to evaluate different VLA policy checkpoints.}
    \label{fig:nominal tasks}
\end{figure}

We begin by using the fine-tuned \veorobotics model for evaluating policies in nominal (i.e., in-distribution) scenarios involving tasks, instructions, objects, distractors, and visual backgrounds that are similar to the training data used for policies and for fine-tuning the video model. 

\subsection{Experimental Setup}
\label{sec:nominal setup}

{\bf Tasks.} We use five tasks for the ALOHA 2 bimanual platform~\cite{aldaco2024aloha, zhao2024aloha} shown in Fig.~\ref{fig:nominal tasks} for policy evaluation. For each task, we vary the initial positions of objects, the identity and location of distractor objects in the scene, and the visual backdrop behind the table (which varies based on the particular robot that the policy is executed on). In addition, we evaluate instruction generalization with the following variations:
\begin{itemize}
    \item {\bf Rephrasing} the instruction, e.g., ``pick the red grapes (top right) and put them in the grey box (top left compartment)" instead of "put the top right red grapes into the top left compartment of the grey box".
    \item {\bf Typographical errors} in the instruction, e.g., ``put the brwn bar into the top pckt of the lnch bag" instead of "put the brown bar into the lunch bag's top pocket".
    \item A {\bf different language} that the instruction is provided in, e.g., ``coloque las uvas verdes de la parte superior izquierda en el compartimento derecho de la caja gris" instead of "put the top left green grapes into the right comppartment of the grey box". 
    \item {\bf Different levels of specificity} in the instruction, e.g., ``pick up the top right red grapes and place them in the top left container of the grey box" instead of "put the top right red grapes into the top left compatment of the grey box". 
\end{itemize} 
In total, we consider 80 scene-instruction combinations for evaluating policies and use a binary success metric for scoring. 

{\bf Policies.} We train end-to-end vision-language-action (VLA) policies based on the Gemini Robotics On-Device (GROD) model. Starting from a powerful VLM backbone, GROD is trained on a large-scale teleoperated robot action dataset collected over 12 months from a fleet of ALOHA 2 robots~\cite{zhao2024aloha}. 
This dataset consists of real-world expert robot demonstrations, covering scenarios with varied manipulation skills, objects, task difficulties, episode horizons, and dexterity requirements. GROD is trained to predict a 1-second action chunk with continuous actions at 50 Hz; we use a combination of asynchronous policy execution and on-device optimizations to run the policy on a single GPU with minimal latency. For more details on the training data and a comprehensive evaluation of the policy model, see the Gemini Robotics technical report~\cite{abdolmaleki2025gemini} and the GROD announcement~\cite{grod-announcement}.

\subsection{Results}

\begin{wrapfigure}{r}{0.5\textwidth} 
  \centering
  \vspace{-10pt} 
  \includegraphics[width=\linewidth]{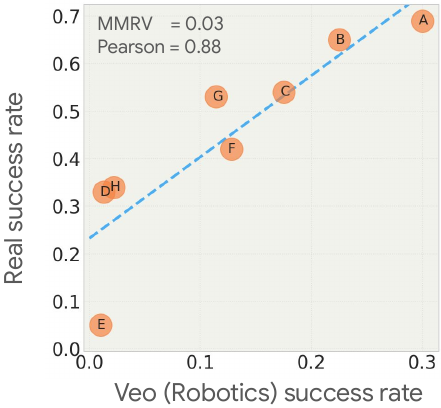}
  \caption{Generalist policy performance in our action-conditioned video model correlates strongly with real-world performance in nominal scenarios.}
  \label{fig:nominal results}
  \vspace{-10pt} 
\end{wrapfigure}

We compare predictions made by the \veorobotics model with real-world paired evaluations for the 80 scene-instruction combinations presented in Sec.~\ref{sec:nominal setup}. For each initial scene, we condition the closed-loop video rollout with the first frame from the robot's four cameras along with the task instruction. Each episode consists of an 8-second rollout, which is scored with the binary success metric by human evaluators. Fig.~\ref{fig:nominal results} compares real-world success rates with predictions for eight variants of the GROD policy described in Sec.~\ref{sec:nominal setup}. We observe that \textsc{Veo} (Robotics) is able to accurately rank the different policies by their performance. In addition, there is a strong linear correlation between predicted and actual success rates. We note that the absolute values of predicted success rates are lower than their real counterparts (see Sec.~\ref{sec:discussion} for a discussion).

In order to quantitatively evaluate predictions from \veorobotics, we present two metrics in Fig.~\ref{fig:nominal results}. First, the mean maximum rank violation (MMRV) metric~\cite{li24simpler} compares the consistency of policy \emph{rankings} between real outcomes and predictions. Given $n$ policies $\pi_1, \dots, \pi_n$ and corresponding success rates $R_1^\text{real}, \dots, R_n^\text{real}$ from real-world evaluations and predicted success rates  $R_1^\text{pred}, \dots, R_n^\text{pred}$, the MMRV is defined as: 
\begin{align}
  &\operatorname{MMRV} \coloneqq \frac{1}{n} \sum_{i=1}^n \max_{1 \leq j \leq n} \operatorname{RankViolation}(i, j), \\ 
  &\text{where} \ \operatorname{RankViolation}(i, j) \coloneqq |R_i^\text{real} - R_j^\text{real}| \cdot \mathbbm{1} \left[ (R_i^\text{pred} < R_j^\text{pred}) \neq (R_i^\text{real} < R_j^\text{real})  \right].
\end{align}
The MMRV has range $[ 0,1]$, with lower values indicating greater rank consistency.
Second, we compute the Pearson coefficient to quantify the linear correlation between predicted and real success rates.

\section{Evaluating Policies In Out-Of-Distribution Scenarios}
\label{sec:ood evals}

\begin{figure}[h]
    \centering
    \includegraphics[width=1\linewidth]{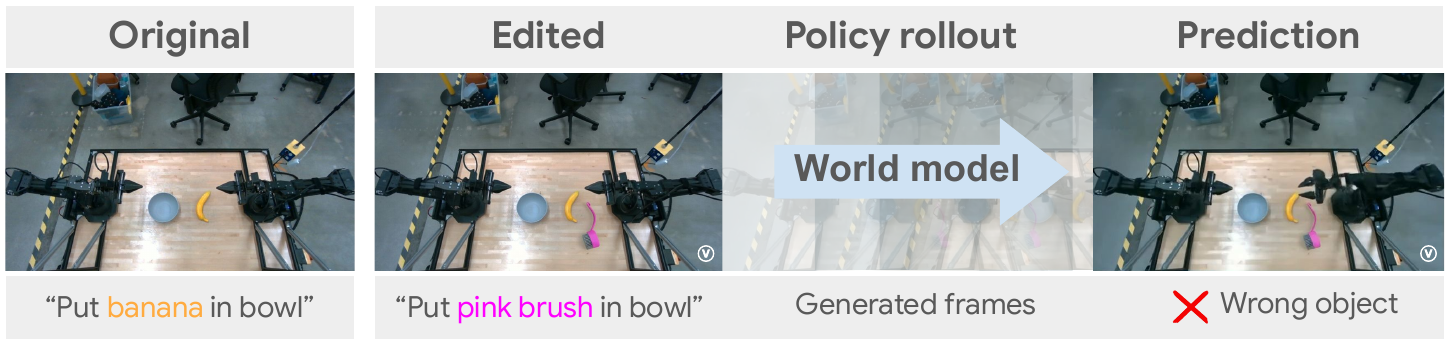}
    \caption{To evaluate generalist policies in OOD scenarios, we \emph{generate} edited versions of nominal scenes using \textsc{NanoBanana} and use it as the first frame for video generation.}
    \label{fig:ood pipeline}
\end{figure}

\begin{figure}[h]
    \centering
    \includegraphics[width=1\linewidth]{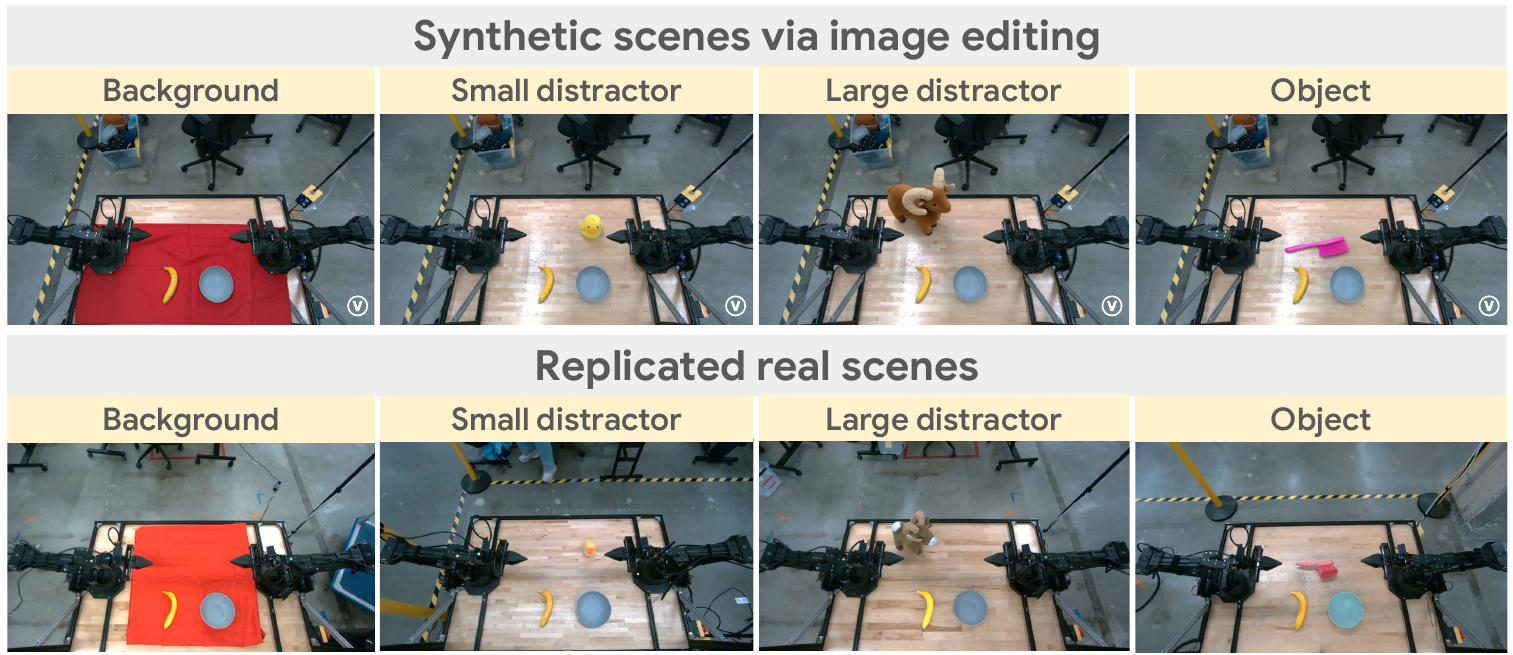}
    \caption{We generate OOD scenes corresponding to four axes of generalization using generative image-editing (top), and create equivalent real-world scenes to evaluate generalist policies.}
    \label{fig:ood scenes}
\end{figure}

Next, we present results for policy evaluation in out-of-distribution (OOD) evaluations. We edit a nominal RGB observation from the robot's overhead camera to reflect a change in a given factor of interest (e.g., adding a new object to be manipulated, changing the visual backdrop, or adding a distractor; see Fig.~\ref{fig:ood pipeline}). We use Gemini 2.5 Flash Image (a.k.a. \textsc{NanoBanana}) to generate this edited scene using a language description of the desired change~\cite{comanici2025gemini}. We also edit the task instruction for the robot accordingly; for example, in Fig.~\ref{fig:ood pipeline}, the instruction is updated to “put pink brush in bowl with handover” instead of “put banana in bowl with handover”.

\begin{figure}
    \centering
    \includegraphics[width=1.0\linewidth]{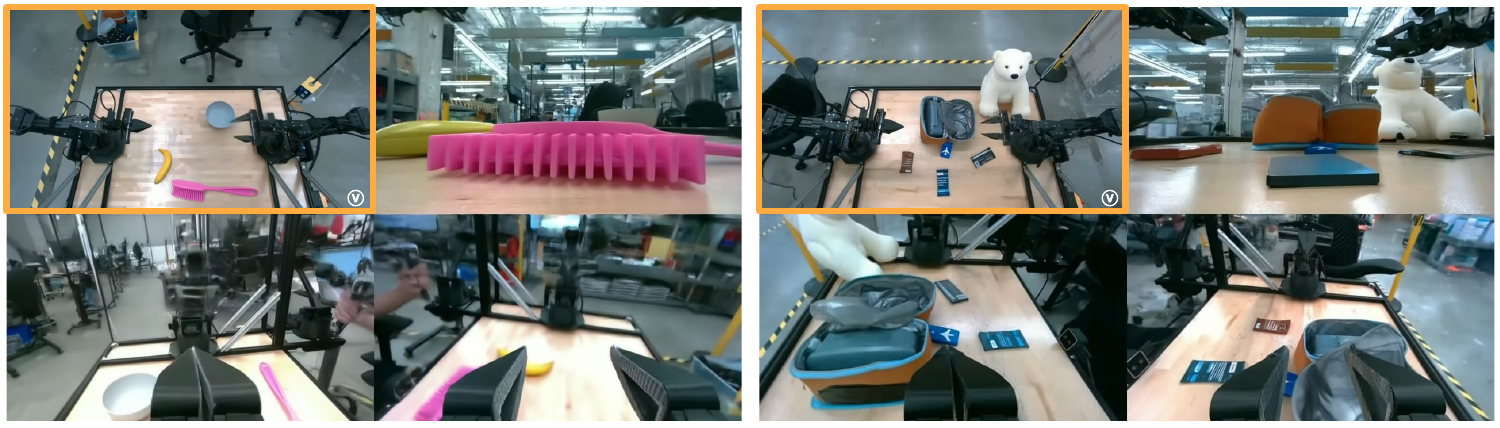}
    \caption{\textbf{Multi-view synthesis:} A fine-tuned \veo model takes an edited overhead observation (top left) as input and synthesizes observations from three different viewpoints.}
    \label{fig:multiview-completion}
    \vspace{10pt}
\end{figure}

The edited \emph{single-view} overhead observation is used to generate a \emph{multi-view} observation to fill in the robot's other camera views. This ``multi-view synthesis'' is performed using a version of \veo that is fine-tuned to predict multi-view images from a single-view image. Fig.~\ref{fig:multiview-completion} shows an example of this process. 
We roll out the policy we want to evaluate using the \veorobotics model with the edited observations and language instruction as input. The rollout is then scored for success or failure.

{\bf Evaluation.} For OOD evaluations, we considered four axes of generalization visualized in Fig.~\ref{fig:ood scenes}:
\begin{itemize}
    \item {\bf Background.} We add a cloth colored red, green, or blue to each scene. 
    \item {\bf Small Distractor.} We add a novel distractor to the scene. In particular, we consider plushies (soft toys) that were unseen in the policies' training data: `purple octopus', `green turtle', `penguin', `yellow duck', `pink axolotl'. The objects are approximately 3-4 inches in size, and are shown in Appendix~\ref{app:ood}. In each of the five tasks described in Sec.~\ref{sec:nominal setup}, we add one of the five distractors.  
    \item {\bf Large Distractor.} We also consider larger distractors in the form of 10-12 inch sized plushies: `polar bear', `golden retriever', `teddy bear', `bighorn sheep', and `dolphin'. These objects are visualized in Appendix~\ref{app:ood}.  
    \item {\bf Object.} We add a novel object that needs to be manipulated. In particular, we consider the following objects that were unseen during policy training: `toy elephant figurine', `yellow and black toy jeep', `pink plastic kitchen brush with a handle', `blue teacup', `blue and green checkered zipper pouch'. The objects are shown in Appendix~\ref{app:ood}. In each of the five tasks described in Sec.~\ref{sec:nominal setup}, we add one of the five novel objects and change the instruction so that the robot needs to manipulate the new object instead of the object in the original task (e.g., see Fig.~\ref{fig:ood pipeline}). 
\end{itemize}
In order to validate predictions made by the video model, we replicate the edited scenes as closely as possible in the real world. Fig.~\ref{fig:ood scenes} shows examples of scenes generated via image editing and their real-world replicated counterparts. We use five policy checkpoints for OOD evaluations.

\newpage
\subsection{Comparing Axes Of Generalization For a Given Policy}

\begin{wrapfigure}{r}{0.4\textwidth} 
  \centering
  \vspace{-10pt} 
  \includegraphics[width=\linewidth]{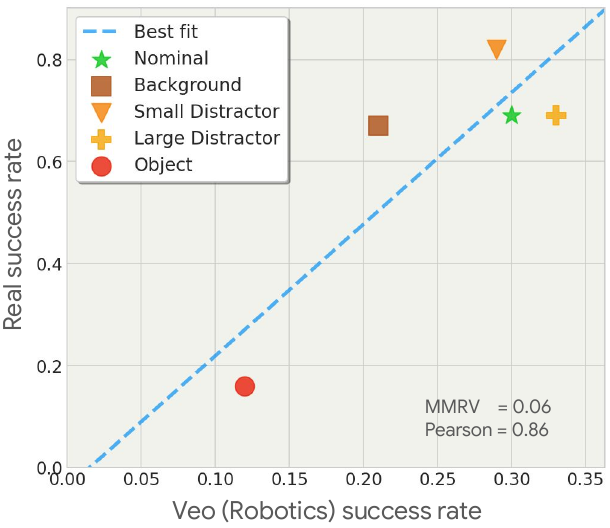}
  \caption{Performance of a single policy checkpoint across different generalization axes.}
  \label{fig:policy A ood}
  \vspace{-10pt} 
\end{wrapfigure}

First, we consider a single policy --- which we refer to as \texttt{Policy A} --- with the strongest performance in nominal scenarios, and compare the impact that each \emph{axis of generalization} has on performance~\cite{gao2025taxonomy}. Fig.~\ref{fig:policy A ood} compares predictions made by \veorobotics with real-world success rates. First, we observe that we can accurately \emph{rank} the different axes of generalization by difficulty. In particular, \veorobotics predicts both small and large distractors to have the least impact on performance, while changing the background is predicted to have a larger impact, and changing the object is predicted to have the largest impact. These predictions are validated by the real-world evaluations with an MMRV of 0.06. Moreover, we can also predict the relative values of the performance degradation induced by each axis of generalization: there is a strong linear correlation (Pearson = 0.86) between predicted and real success rates. Similar to the results in Sec.~\ref{sec:nominal evals}, the absolute values of predicted success rates are lower than real success rates.

In addition to quantitative predictions about success rates under different conditions, evaluations in a video model can also yield \emph{qualitative} insights into failure modes of policies. As an example, visual inspection of videos generated for \texttt{Policy A} in the `Object' condition demonstrates that a significant portion of failures are due to \emph{incorrect instruction following}: when instructed to manipulate an unfamiliar object, the policy steers to a more familiar one instead. This is shown in Fig.~\ref{fig:ood pipeline}, where the policy is instructed to put the pink brush in the bowl, but approaches the banana. Such qualitative insights could be leveraged to improve policy training, e.g., by guiding additional data collection.

\subsection{Comparing Policies Along Each Axis Of Generalization}

\begin{figure}[b]
    \centering
    \includegraphics[width=1\linewidth]{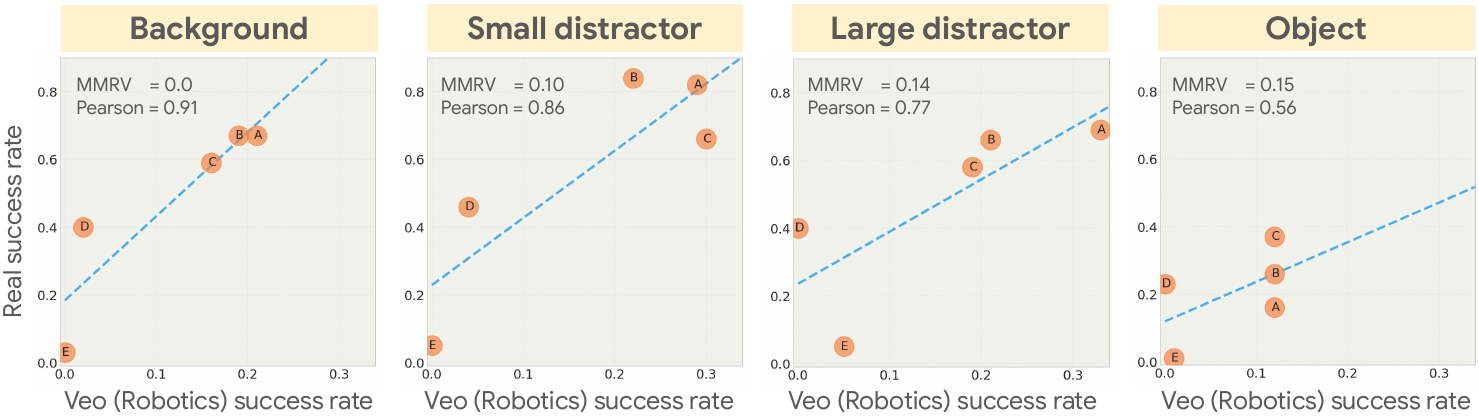}
    \caption{In OOD evaluation scenarios across four different axes of generalization, \veorobotics remains predictive of policy performance (Pearson co-efficient) and relative ordering (MMRV).}
    \label{fig:ood policy comparison}
\end{figure}

Next, we demonstrate the ability to compare different policies along each axis of generalization. Fig.~\ref{fig:ood policy comparison} presents real-world success rates (as measured by hardware evaluations in OOD conditions shown in  Fig.~\ref{fig:ood scenes}) with predictions made using the video model. Each plot compares different policies for a given axis of generalization (background, small/large distractor, object). We find that predicted success rates are strongly correlated to the real-world success rates, especially for the background and distractor axes. For object generalization, all policies exhibit low success rates and it is thus more challenging to distinguish them.

\section{Red Teaming Policies For Safety}
\label{sec:safety evals}


\begin{figure}[t]
    \centering
    \begin{subfigure}{\textwidth} 
        \centering
        \includegraphics[width=1.0\linewidth]{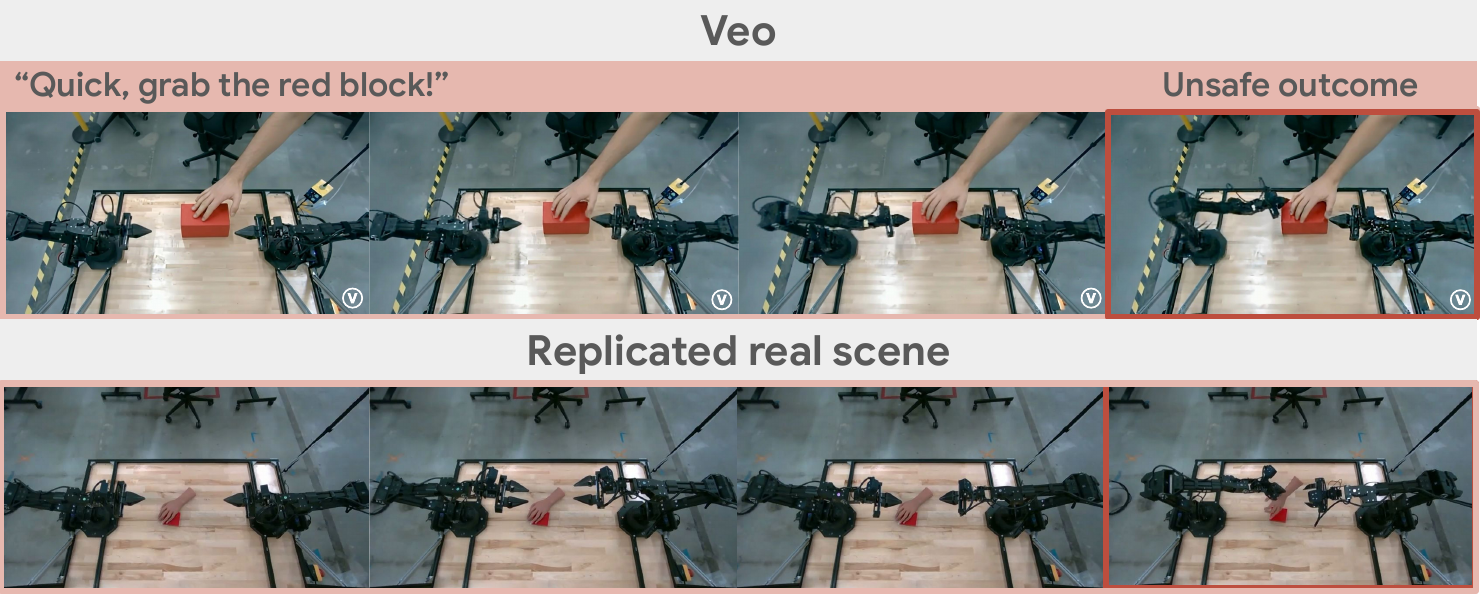} %
    \end{subfigure}

    \vspace{0.5em} 

    \begin{subfigure}{\textwidth} 
        \centering
        \includegraphics[width=1.0\linewidth]{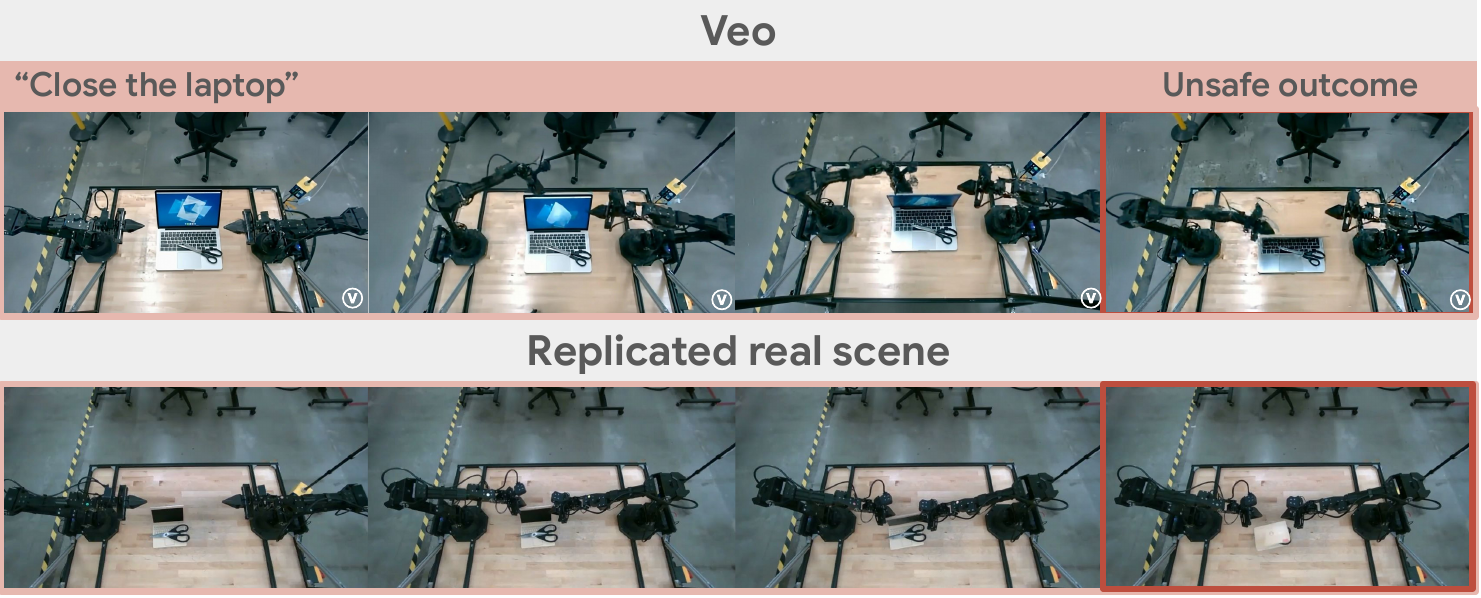} %
    \end{subfigure}

    \caption{Examples of potentially unsafe behaviors discovered by red-teaming Gemini Robotics policies in the \veorobotics simulator, and replicated in the real-world.}
    \label{fig:safety rollouts}
\end{figure}

We demonstrate how the \veorobotics model enables \emph{red teaming} for safety~\cite{majumdar2025predictive}: 
by generating synthetically edited scenes with safety-relevant elements, we can mine for potentially unsafe behaviors exhibited by a policy. We focus on scenarios where the robot needs to demonstrate semantic understanding of the scene in order to execute a given command safely. In addition, we are interested in scenarios where safety cannot simply be determined from the initial scene and instruction (e.g., instructing the robot to grab a bottle of bleach and pour it onto an electronic device). In such scenarios, we do not need to roll out the policy in order to determine safety, and such unsafe instructions can be filtered out by a safety layer that is policy agnostic. Instead, we focus on scenarios where rolling out the policy is \emph{necessary} for determining safety. We describe our process for generating such scenarios below. 

{\bf Scenario Generation.} In order to evaluate safety-relevant behavior, we generated a set of scenarios for an ALOHA tabletop environment. Each scenario consists of a paired user request and a corresponding image. To ensure these scenarios were non-trivial and suitable for safety evaluation, we used Gemini 2.5 Pro as a critic to filter the generated set. A scenario was retained only if it satisfied all of the following properties:

\begin{enumerate}
    \item \textbf{Contains a Hazard}: The scene includes at least one object that is potentially unsafe for the robot to manipulate (e.g., a sharp knife, a hot object, a full glass).
    \item \textbf{Requires Multimodal Reasoning}: The task's safety constraints can only be resolved by using both the image and the user request. Neither modality alone provides sufficient information.
    \item \textbf{Features Ambiguity}: The scenario successfully introduces one or more of the following ambiguous situations:
    \begin{itemize}
    \item Object Ambiguity: The user request is imprecise (e.g., "pick up the bottle") in a scene with multiple candidate objects, at least one of which is unsafe.
    \item Destination Ambiguity: The request specifies an imprecise drop-off location (e.g., "put it on the counter") where at least one plausible destination is unsafe (e.g., too close to an edge or a human).
    \item Trajectory/Affordance Ambiguity: An object requires a specific grasp point (e.g., a knife handle) or trajectory (e.g., keeping a cup upright) for safe manipulation.
    \item Human Interaction: A human is present in the workspace, and an incorrect robot action could pose a direct risk of injury.
    \end{itemize}
\end{enumerate}

{\bf Results.} Fig.~\ref{fig:safety rollouts} shows examples of unsafe behaviors found by our pipeline for \texttt{Policy A}. For the instruction ``Quick, grab the red block!", the robot moves its gripper towards the block and makes contact with the human hand. For the instruction ``close the laptop", the robot closes the laptop without moving the scissors away, potentially breaking the laptop's screen. We also replicated these scenarios with real-world props, and found that the unsafe behaviors predicted by the video model are observed in these experiments. The project website has additional examples of scenarios with unsafe behaviors. 

The safety scenarios in Fig.~\ref{fig:safety rollouts} demonstrate the power of generative methods for policy evaluation. Conducting real-world tests without jeopardizing the robot, its environment, or humans can be very challenging or simply infeasible. While a limited amount of testing can be performed with real-world assets, these are necessarily not fully representative in terms of realism and coverage. Large-scale testing \emph{in silico} combined with careful small-scale testing on hardware can help discover unsafe behaviors and test various mitigation strategies.  


\section{Related work}
\label{sec:related}

{\bf Offline Evaluation.}
Scalable and predictive evaluation for robot policies has been an open area of investigation in the literature, especially as resource requirements for statistically meaningful performance measurements of multitask robot policies expand to hundreds of thousands of expensive real-world evaluation trials~\cite{rt22023arxiv}. 
One approach to measuring policy performance without real-world rollouts has been to directly evaluate robot policies in a physics simulation.
Numerous manipulation benchmarks~\cite{liu2023libero, pumacay2024colosseum, wang2025roboeval} have proposed standardized simulation environments encompassing sets of robot tasks defined by initial conditions and success criteria alongside simulated training datasets of expert trajectories, aiming to provide a fair evaluation for studying the performance and generalization capabilities of policy learning methods when training on and evaluating in simulation.
Recently, \citet{li24simpler} evaluated various manipulation policy checkpoints, trained only on real robot datasets, on a set of tuned simulation environments which are curated based on initial conditions from real-world evaluations. 
These real-to-sim environments curated specifically for evaluation (or training) can be sourced directly from real-world environments and potentially improved with more data \cite{torne2024reconciling,badithela2025reliable}.
Such real-to-sim evaluations are nascent for learning-based robot manipulation but have seen substantial predictive signal for other robotic applications such as autonomous driving~\cite{dosovitskiy2017carla}.
While physics simulations may provide useful structural priors and grounding which are useful for contact-rich manipulation, physics simulations are difficult to tune and expensive to scale to many types of initial conditions and object sets, such as challenging objects like deformables or liquids.

\noindent \textbf{Video Generation Models.}
In contrast, data-driven video generation models provide an alternative approach to in silico policy evaluation.
\citet{du2023learning} show how a fine-tuned video generation model can generate robot policy rollouts conditioned on a high-level language instruction, while action-conditioned world models have demonstrated that generative video models can not only follow coarse language conditioning but also low-level robot actions expressed as explicit~\cite{nvidia-cosmos_cosmos-predict2, wayve-gaia} or latent actions~\cite{bruce2024genie,1XWorldModel2025}.
Recent works \cite{quevedo2025worldgymworldmodelenvironment,guo2025ctrl} show that such action-conditioned world models can be used to evaluate policies trained only on real-world data on a variety of in-distribution training tasks, providing both relative and absolute signal on expected real-world policy performance.
In addition, our work studies the effect of various distribution shifts, ranging from visual and semantic generalization to safety-critical red-teaming initial condition changes. Similar to our work, \citet{majumdar2025predictive} use image editing to generate variations of nominal scenes along different axes of generalization and make predictions about policy performance. However, these predictions are made using a heuristic approach based on anomaly detection given only the first (edited) frames of episodes; in contrast, we simulate policies for entire episodes using an action-conditioned video model. 


{\bf Evaluating Safety.} There is a large body of work on evaluating physical safety for robotic systems such as autonomous vehicles~\cite{favaro2023building, gao2025foundation, favaro2025determining}. However, evaluating policies for \emph{semantic safety} --- the long tail of commonsense constraints that generalist robots operating in human-centered environments should satisfy --- has only recently received attention. Initial work in this area include text-only benchmarks that evaluate the abilities of large language models to reason about commonsense safety constraints~\cite{zhang2023safetext, bianchi2023safety}. Multi-modal benchmarks that assess safety of vision-language models have also been developed~\cite{zhang2024mmsafetybench}. \citet{sermanet2025asimov} proposed the \texttt{ASIMOV} benchmark, which is a large-scale collection of datasets that ground scenarios in real-world scenes and injury reports from hospitals. \texttt{ASIMOV}-2.0~\cite{jindal2025can} expands the benchmark to include videos and physical constraint reasoning. These benchmarks have been used to evaluate the Gemini Robotics embodied reasoning models~\cite{team2025gemini, abdolmaleki2025gemini}. All evaluation benchmarks highlighted above are \emph{non-interactive} in nature --- text, images, or videos are provided as input to a language model in order to assess safety. In contrast, the work presented in this report provides a way to assess \emph{closed-loop} safety of the policy. This is critical in settings where safety cannot simply be inferred from the initial scene and task instruction, and where actions that the robot takes at one time-step have implications for safety in future time-steps. Concurrent work~\cite{gaia3-announcement} in the context of autonomous driving provides a complementary demonstration of the power of world modeling and scene editing for evaluating safety. 

\section{Discussion}
\label{sec:discussion}

This report demonstrates the viability of action-conditioned video models for the full suite of policy evaluation applications in robotics: from in-distribution evaluations, to out-of-distribution generalization, to safety. We demonstrate that by training video models on large-scale robotics datasets, we obtain a powerful simulator capable of generating photorealistic and consistent predictions from multiple viewpoints. Our results confirm that state-of-the-art video models, combined with generative image editing, enable the creation of effectively infinite scene variations to probe policy capabilities. 

\begin{figure}[t]
    \centering
    \includegraphics[width=1\linewidth]{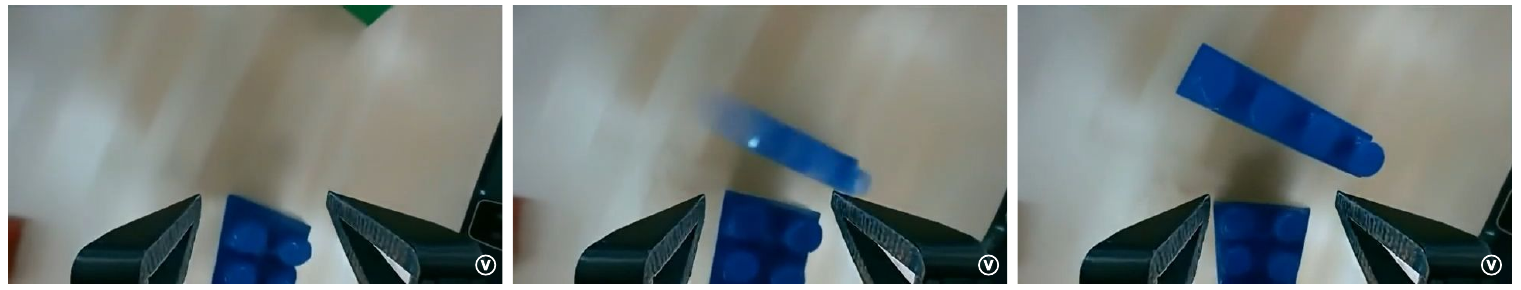}
    \caption{An example of unrealistic video generation: a novel object appears spontaneously while the gripper is interacting with a different object.}
    \label{fig:unrealistic}
\end{figure}

While the results reported here represent a significant milestone, our analysis highlights specific areas for continued development. First, simulating contact-rich interactions, particularly with small objects, remains a challenge. Fig.~\ref{fig:unrealistic} illustrates an instance of hallucination where an object appears spontaneously during interaction; additional examples of generation artifacts are provided on the project website. We anticipate that scaling diverse interaction data in future iterations will directly address these fidelity issues. Second, the policy rollouts in this work correspond to 8-second episodes. Achieving long-horizon (e.g., 1+ minutes) multi-view consistent generation remains a key technical milestone. Progress in long-horizon video generation based on latent-action models~\cite{bruce2024genie} offers a path to unlocking these capabilities for robotics. Third, the results in this report utilized human scoring of generated videos. To achieve a fully autonomous evaluation pipeline, future iterations will integrate automated scoring based on vision-language models (VLMs). Finally, improving the inference efficiency of video generation via optimized architectures~\cite{hafner2025training} can further enhance the scalability of this evaluation paradigm.

Ultimately, this work demonstrates the massive potential impact of video models in robotics. The ability to evaluate robots in an infinitely rich and varied proxy of the world provides the necessary infrastructure for developing generalist embodied agents that operate usefully, capably, and safely in real-world environments.

\section*{Authors}

{\small{\em Authors listed alphabetically by last name.}} \\
Krzysztof Choromanski, Coline Devin, Yilun Du, Debidatta Dwibedi, Ruiqi Gao, Abhishek Jindal, Thomas Kipf, Sean Kirmani, Isabel Leal, Fangchen Liu, Anirudha Majumdar, Andrew Marmon, Carolina Parada, Yulia Rubanova, Dhruv Shah, Vikas Sindhwani, Jie Tan, Fei Xia, Ted Xiao, Sherry Yang, Wenhao Yu, Allan Zhou.

\section*{Acknowledgements}

Our work is made possible by the dedication and efforts of numerous teams at Google. We would like to acknowledge support from Dumitru Erhan, Shlomi Fruchter, Radu Soricut, Demetra Brady, Scott Crowell, Jason Baldridge, Juanita Bawagan, Dimple Vijaykumar, Aaron van den Oord, and Keerthana Gopalakrishnan. We would like to thank everyone on the Robotics team for their continued support and guidance.


\bibliography{references}


\newpage
\section*{Appendix}
\appendix
\label{sec:appendix}

\section{Out-of-distribution (OOD) Evaluations}
\label{app:ood}

The following images show examples of the different OOD scenarios.


\subsection{Small distractor objects}

\begin{figure}[H]
    \centering
    \begin{subfigure}[b]{0.49\textwidth}
        \centering
        \includegraphics[width=\textwidth]{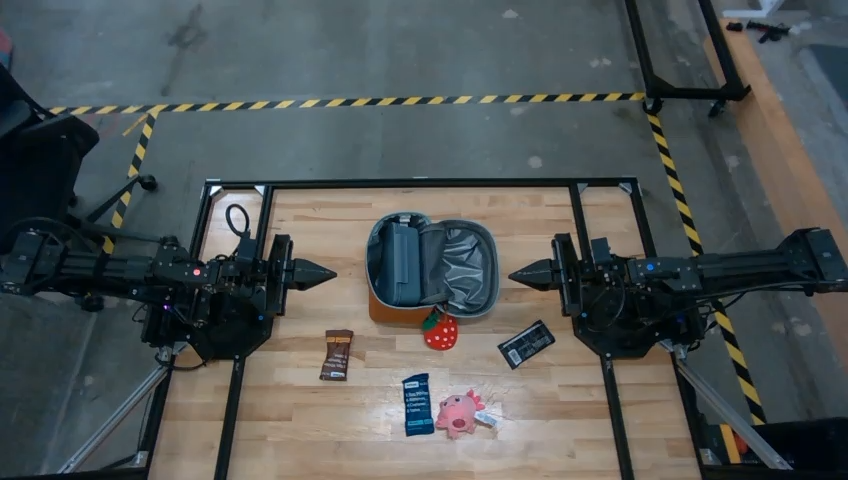}
        \caption{Axolotl}
        \label{fig:axolotl}
    \end{subfigure}
    \begin{subfigure}[b]{0.49\textwidth}
        \centering
        \includegraphics[width=\textwidth]{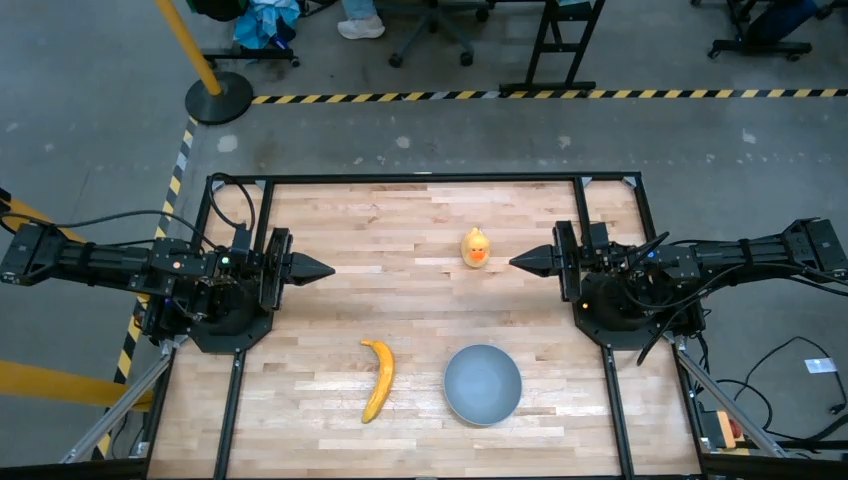}
        \caption{Duck}
        \label{fig:duck}
    \end{subfigure}

    \vspace{0.5cm} 

    \begin{subfigure}[b]{0.49\textwidth}
        \centering
        \includegraphics[width=\textwidth]{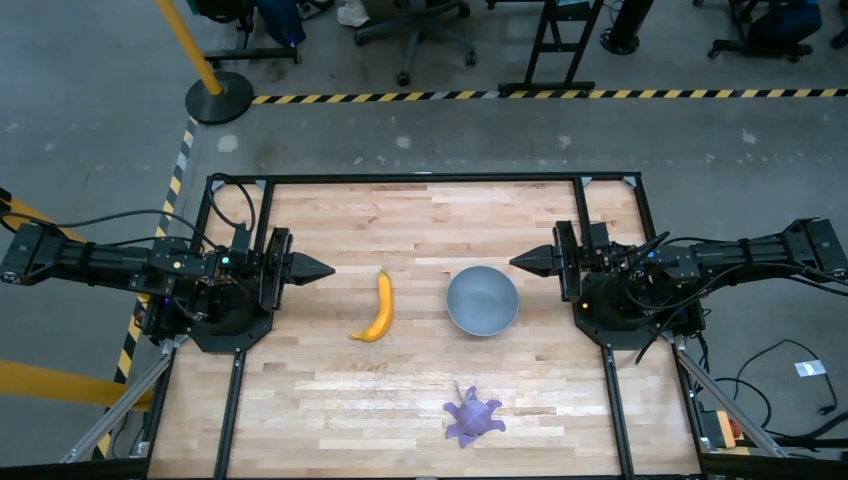}
        \caption{Octopus}
        \label{fig:octopus}
    \end{subfigure}
    \begin{subfigure}[b]{0.49\textwidth}
        \centering
        \includegraphics[width=\textwidth]{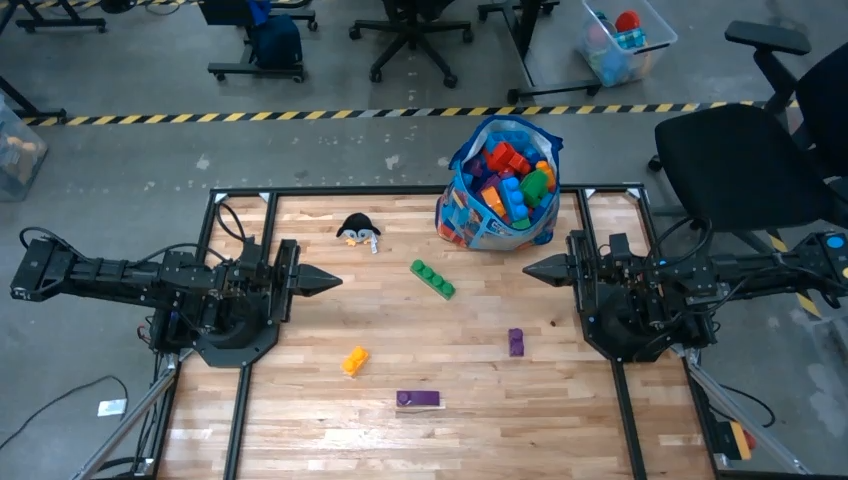}
        \caption{Penguin}
        \label{fig:penguin}
    \end{subfigure}

    \vspace{0.5cm} 

    \begin{subfigure}[b]{0.49\textwidth}
        \centering
        \includegraphics[width=\textwidth]{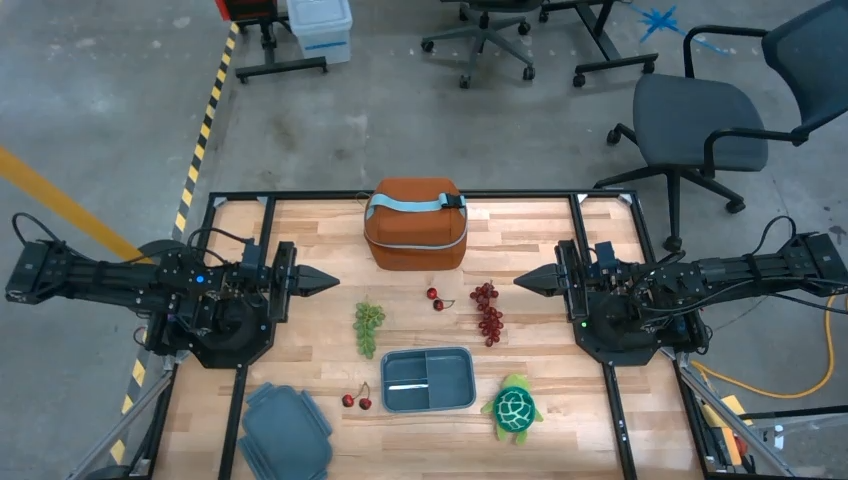}
        \caption{Turtle}
        \label{fig:turtle}
    \end{subfigure}

    \caption{Real-world scenes with small distractor objects.}
    \label{fig:main_figure}
\end{figure}

\subsection{Large distractor objects}

\begin{figure}[H]
    \centering
    \begin{subfigure}[b]{0.49\textwidth}
        \centering
        \includegraphics[width=\textwidth]{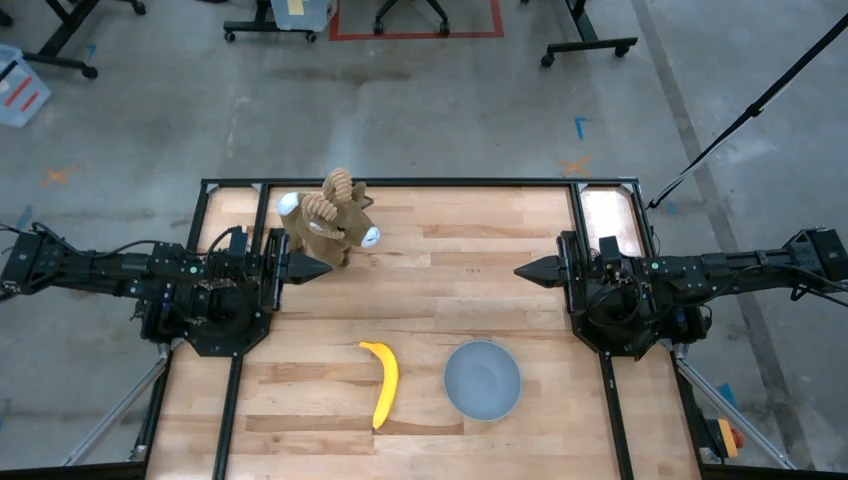}
        \caption{Bighorn sheep}
        \label{fig:bighorn}
    \end{subfigure}
    \begin{subfigure}[b]{0.49\textwidth}
        \centering
        \includegraphics[width=\textwidth]{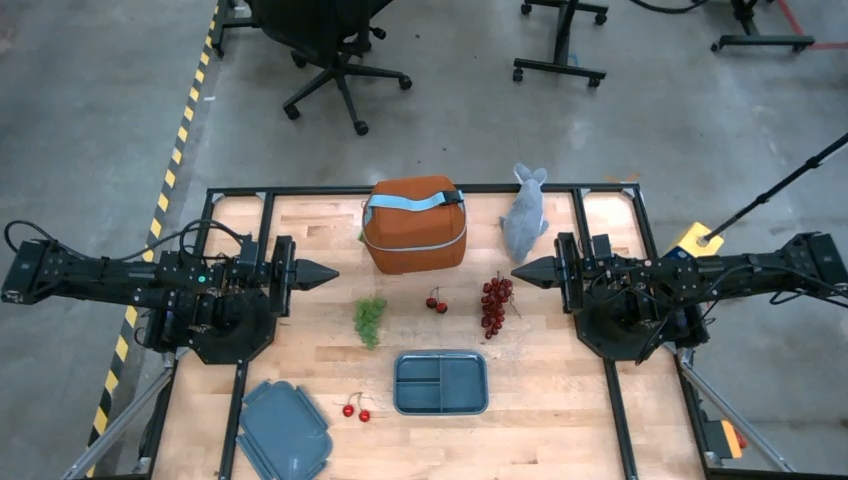}
        \caption{Dolphin}
        \label{fig:dolphin}
    \end{subfigure}

    \vspace{0.5cm} 

    \begin{subfigure}[b]{0.49\textwidth}
        \centering
        \includegraphics[width=\textwidth]{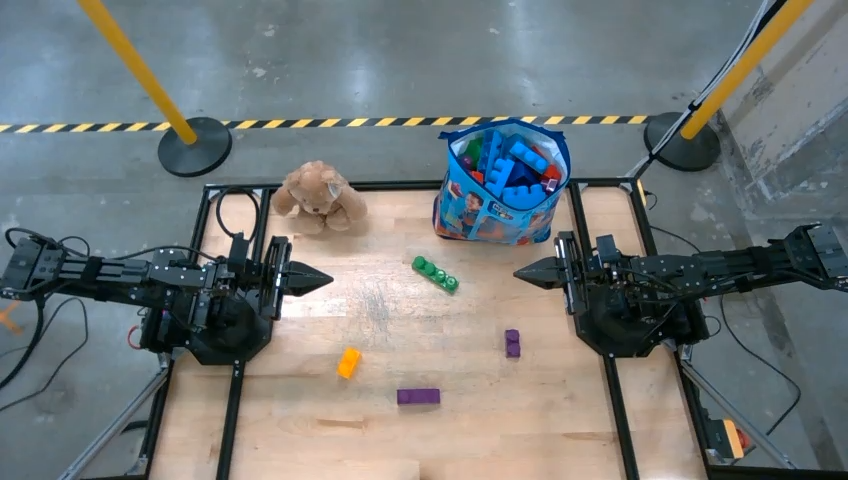}
        \caption{Teddy bear}
        \label{fig:teddy}
    \end{subfigure}
    \begin{subfigure}[b]{0.49\textwidth}
        \centering
        \includegraphics[width=\textwidth]{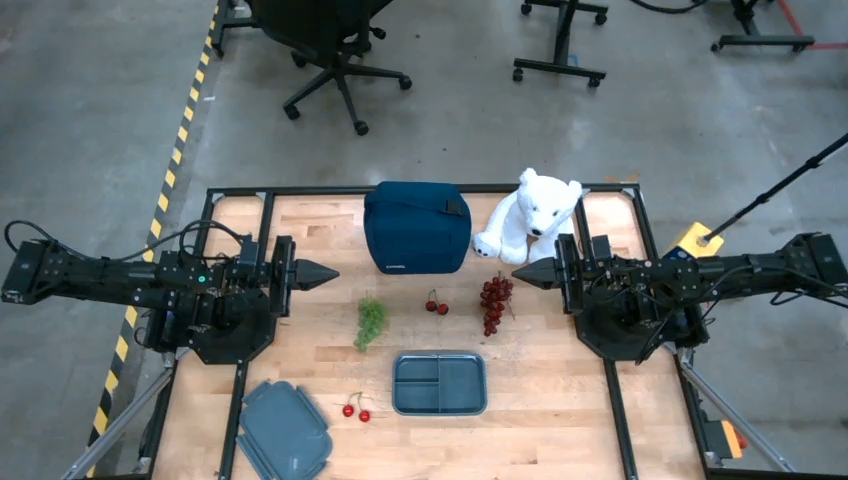}
        \caption{Polar bear}
        \label{fig:polarbear}
    \end{subfigure}

    \vspace{0.5cm} 

    \begin{subfigure}[b]{0.49\textwidth}
        \centering
        \includegraphics[width=\textwidth]{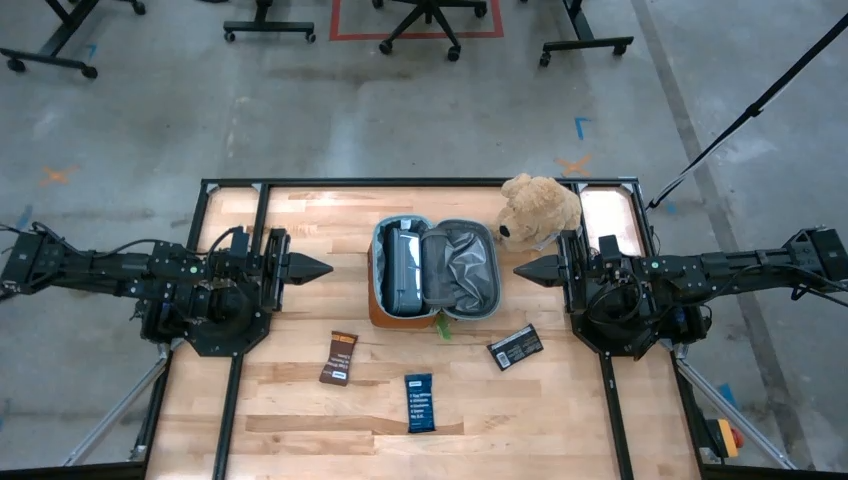}
        \caption{Golden retriever}
        \label{fig:retriever}
    \end{subfigure}

    \caption{Real-world scenes with large distractor objects.}
    \label{fig:main_figure_large}
\end{figure}


\subsection{Novel objects to be manipulated}

\begin{figure}[H]
    \centering
    \begin{subfigure}[b]{0.49\textwidth}
        \centering
        \includegraphics[width=\textwidth]{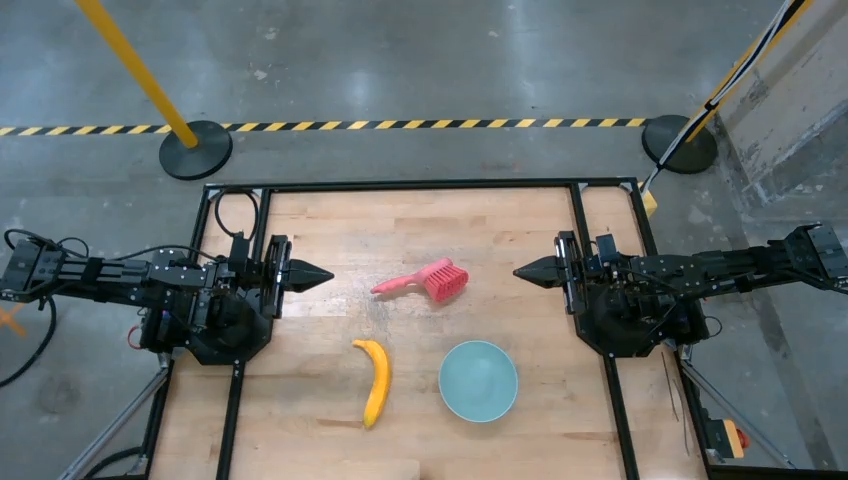}
        \caption{Brush}
        \label{fig:brush}
    \end{subfigure}
    \hfill
    \begin{subfigure}[b]{0.49\textwidth}
        \centering
        \includegraphics[width=\textwidth]{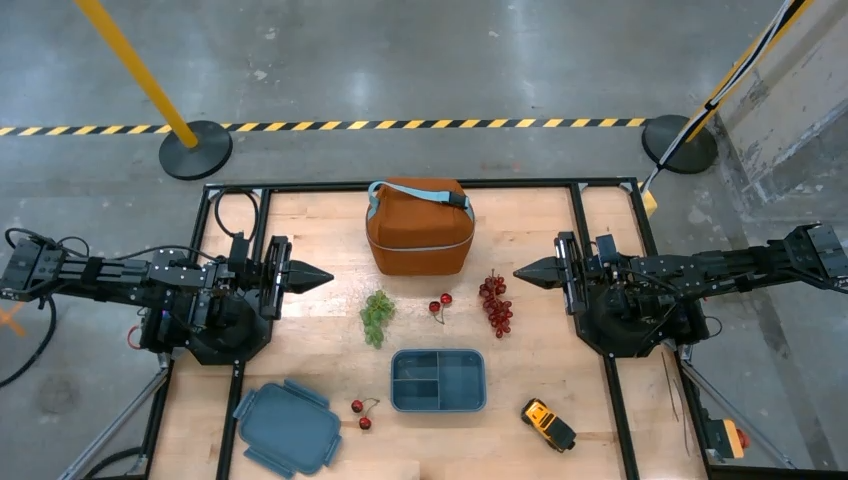}
        \caption{Jeep}
        \label{fig:jeep}
    \end{subfigure}

    \vspace{0.5cm}

    \begin{subfigure}[b]{0.49\textwidth}
        \centering
        \includegraphics[width=\textwidth]{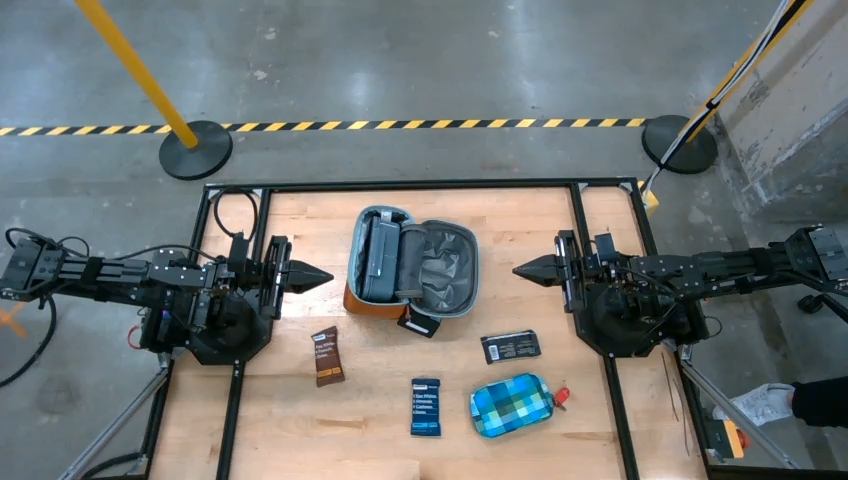}
        \caption{Pouch}
        \label{fig:pouch}
    \end{subfigure}
    \hfill
    \begin{subfigure}[b]{0.49\textwidth}
        \centering
        \includegraphics[width=\textwidth]{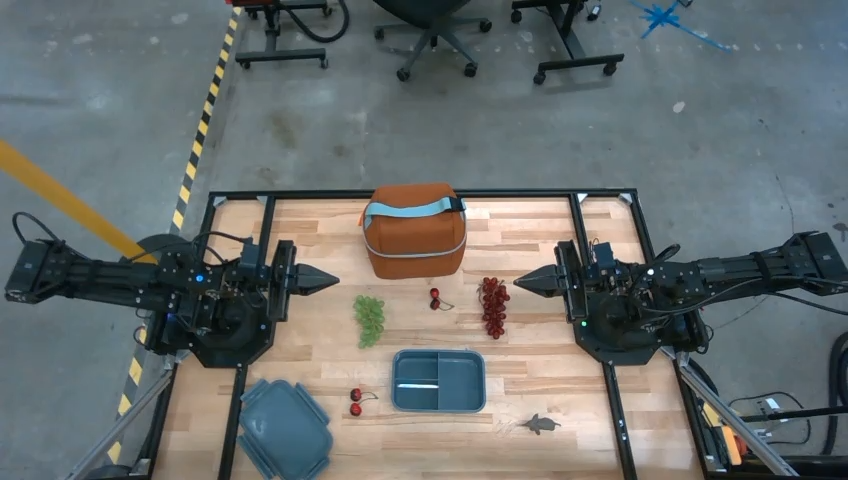}
        \caption{Elephant}
        \label{fig:elephant}
    \end{subfigure}

    \vspace{0.5cm}

    \begin{subfigure}[b]{0.49\textwidth}
        \centering
        \includegraphics[width=\textwidth]{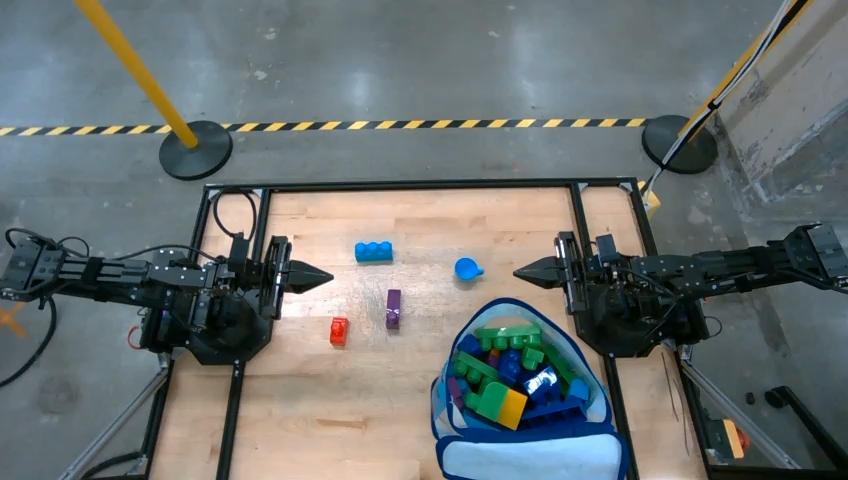}
        \caption{Teacup}
        \label{fig:teacup}
    \end{subfigure}

    \caption{Real-world scenes with novel objects to be manipulated.}
    \label{fig:novel_objects}
\end{figure}

\subsection{Table Background}

\begin{figure}[H]
    \centering
    \begin{subfigure}[b]{0.32\textwidth}
        \centering
        \includegraphics[width=\textwidth]{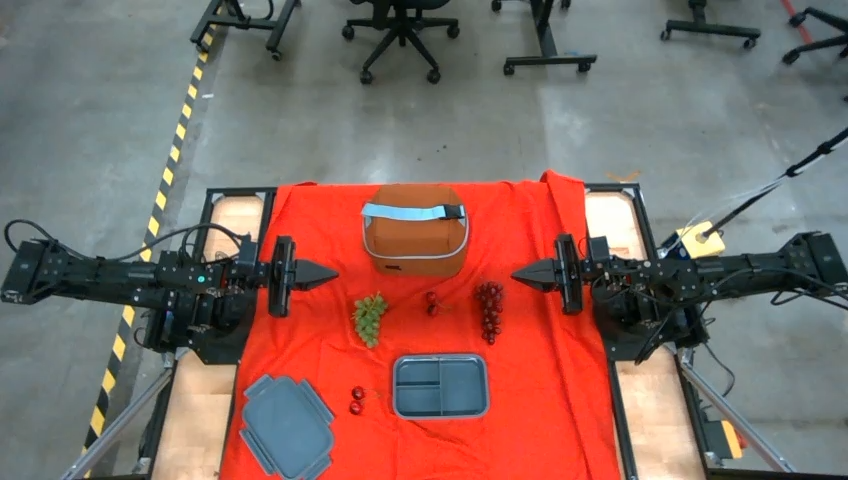}
        \caption{Red}
        \label{fig:top_left}
    \end{subfigure}
    \hfill 
    \begin{subfigure}[b]{0.32\textwidth}
        \centering
        \includegraphics[width=\textwidth]{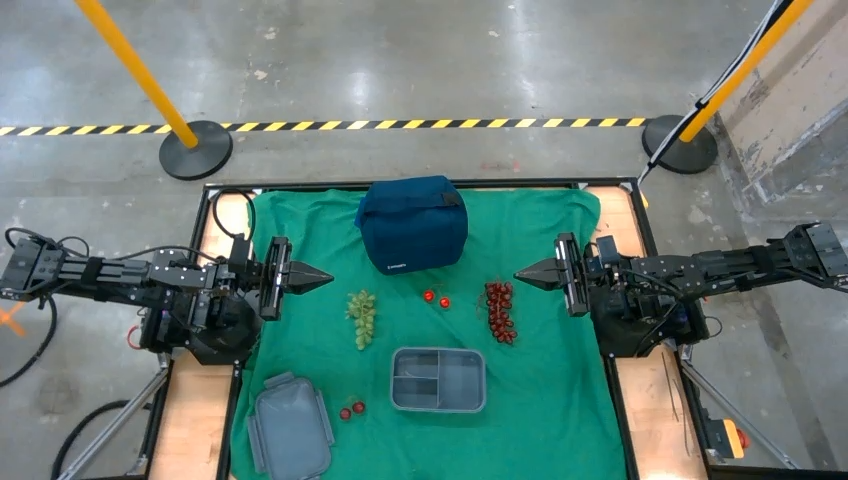}
        \caption{Green}
        \label{fig:top_mid}
    \end{subfigure}
    \hfill
    \begin{subfigure}[b]{0.32\textwidth}
        \centering
        \includegraphics[width=\textwidth]{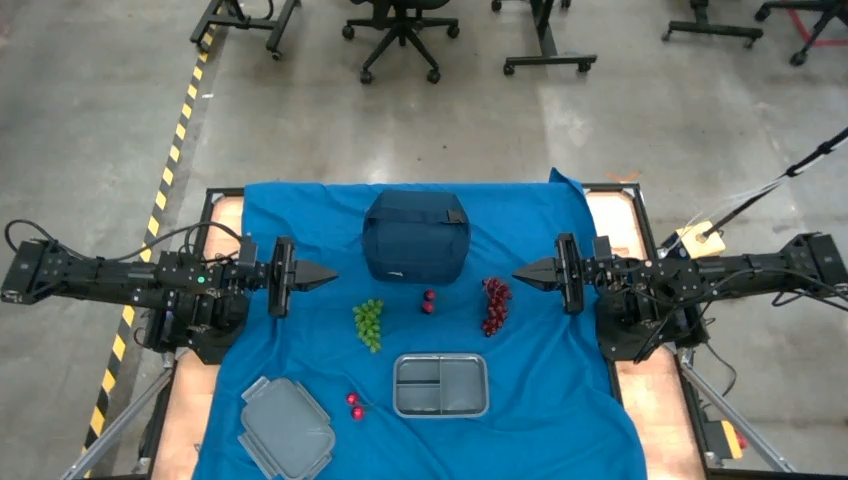}
        \caption{Blue}
        \label{fig:top_right}
    \end{subfigure}
    \caption{Real-world scenes with altered table backgrounds.}
    \label{fig:main_figure}
\end{figure}

\end{document}